\documentclass[10pt,twocolumn,letterpaper]{article}
\usepackage[pagenumbers]{cvpr} 
\usepackage{times}
\usepackage{epsfig}
\usepackage{graphicx}
\usepackage{amsmath}
\usepackage{amssymb}
\usepackage{booktabs} 
\usepackage{soul}
\usepackage{xcolor}

\usepackage{algorithm}
\usepackage{algorithmicx}
\usepackage{algpseudocode}

\DeclareMathOperator*{\argmin}{argmin}

\usepackage[pagebackref=true,breaklinks=true,letterpaper=true,colorlinks,bookmarks=false]{hyperref}

\usepackage[capitalize]{cleveref}
\crefname{section}{Sec.}{Secs.}
\Crefname{section}{Section}{Sections}
\Crefname{table}{Table}{Tables}
\crefname{table}{Tab.}{Tabs.}

\def\confName{CVPR}
\def\confYear{2022}

\begin{document}

\title{Going Grayscale: The Road to Understanding and Improving Unlearnable Examples}

\author{
Zhuoran Liu\\
Radboud University\\
{\tt\small z.liu@cs.ru.nl}

\and
Zhengyu Zhao\\
Radboud University\\
{\tt\small z.zhao@cs.ru.nl}

\and
Alex Kolmus\\
Radboud University\\
{\tt\small alex.kolmus@ru.nl}

\and
Tijn Berns\\
Radboud University\\
{\tt\small tijnberns.berns@ru.nl}

\and
Twan van Laarhoven\\
Radboud University\\
{\tt\small tvanlaarhoven@cs.ru.nl}

\and
Tom Heskes\\
Radboud University\\
{\tt\small tom.heskes@ru.nl}

\and
Martha Larson\\
Radboud University\\
{\tt\small m.larson@cs.ru.nl}

\and

}

\maketitle
 
\begin{abstract}
Recent work has shown that imperceptible perturbations can be applied to craft unlearnable examples (ULEs), i.e. images whose content cannot be used to improve a classifier during training. 
In this paper, we reveal the road that researchers should follow for understanding ULEs and improving ULEs as they were originally formulated (ULEOs).
The paper makes four contributions.
First, we show that ULEOs exploit color and, consequently, their effects can be mitigated by simple grayscale pre-filtering, without resorting to adversarial training.
Second, we propose an extension to ULEOs, which is called ULEO-GrayAugs, that forces the generated ULEs away from channel-wise color perturbations by making use of grayscale knowledge and data augmentations during optimization.
Third, we show that ULEOs generated using Multi-Layer Perceptrons (MLPs) are effective in the case of complex Convolutional Neural Network (CNN) classifiers, suggesting that CNNs suffer specific vulnerability to ULEs.
Fourth, we demonstrate that when a classifier is trained on ULEOs, adversarial training will prevent a drop in accuracy measured both on clean images and on adversarial images.
Taken together, our contributions represent a substantial advance in the state of art of unlearnable examples, but also reveal important characteristics of their behavior that must be better understood in order to achieve further improvements.
Code is available at~\url{https://github.com/liuzrcc/ULE-GrayAug}.

\end{abstract}


\begin{figure}[t]
\centering
\includegraphics[width=0.8\linewidth]{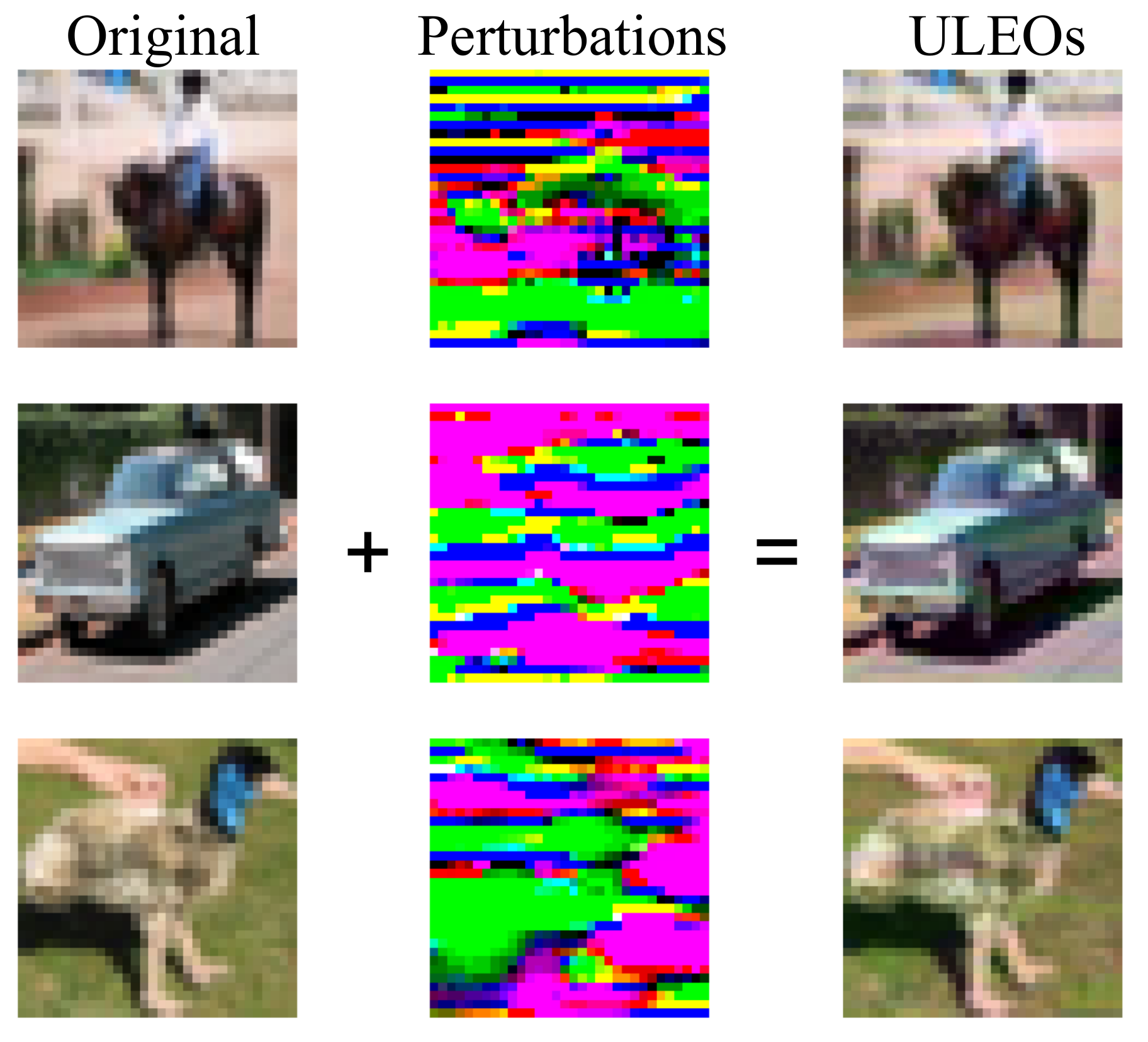}
\caption{Unlearnable examples generated by ULEO~\cite{huang2021unlearnable}.
The imperceptible perturbations ($L_{\infty}$ norm, $\epsilon=8$) contain few spatial but many color channel-wise changes, which inspires our new method of using simple grayscale pre-filtering for defeating ULEOs. Perturbations are multiplied by 255/8 for visualization. More examples can be found in Appendix~\ref{sec:uleoexamples}.}
\label{fig:fig1}
\end{figure}

\begin{figure}[!t]
\centering
\includegraphics[width=0.9\columnwidth]{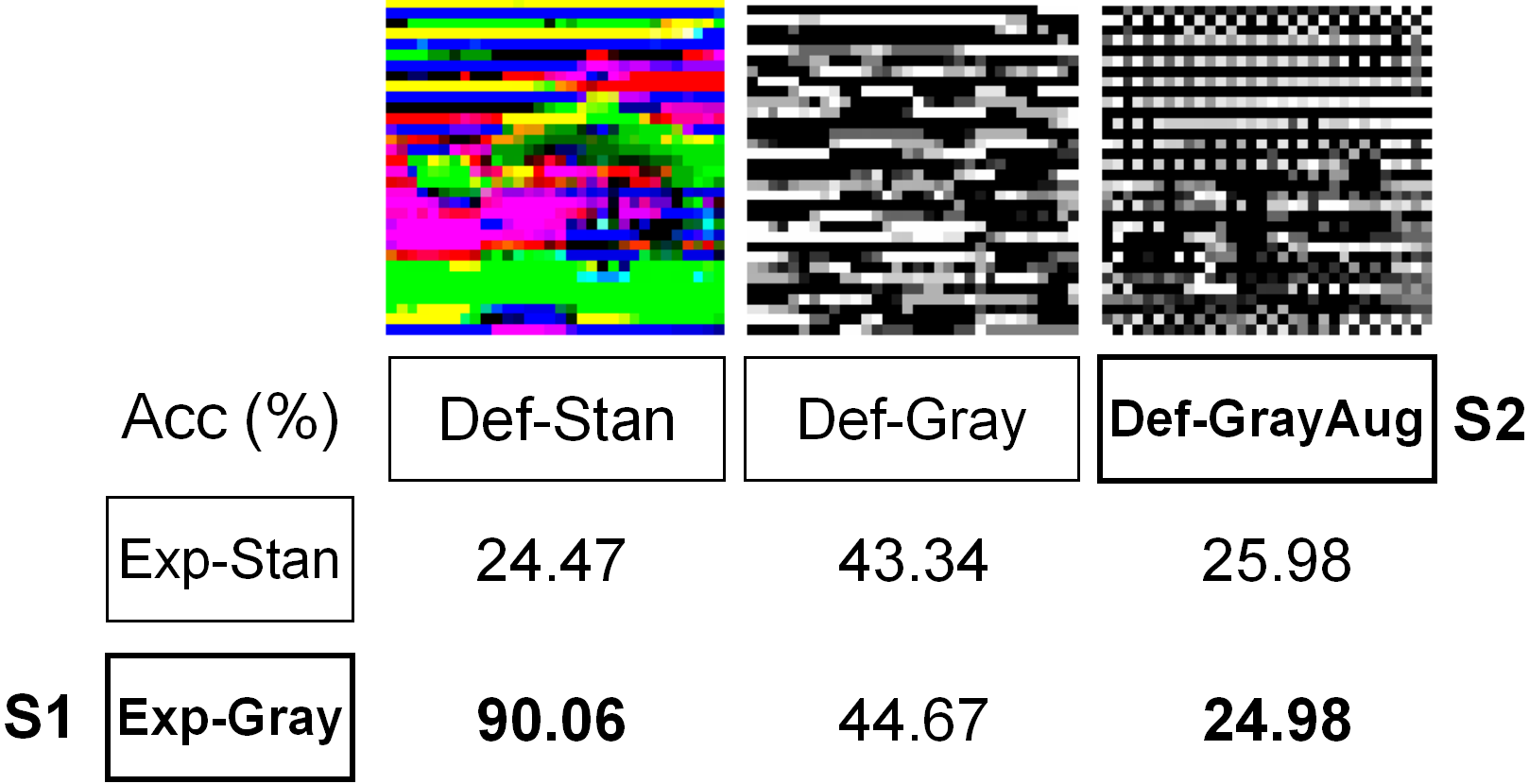}
\caption{
Overview of our work on UnLearnable Examples (ULEs) in two scenarios (see Section~\ref{sec:sce}):
\textbf{S1: Reactive Exploiter}: The exploiter reactively defeats ULEs by suppressing their channel-wise changes via applying grayscale pre-filtering (Exp-Gray) to the standard exploiter (Exp-Stan).
\textbf{S2: Adaptive Defenders}: The standard defender (Def-Stan) is aware of Exp-Gray, and so adapts itself to be Def-Gray, which generates grayscale ULE perturbations, and further to be Def-GrayAug, which additionally promotes spatial perturbations by data augmentations.
Perturbations are for the top image in Figure~\ref{fig:fig1}.
More examples generated by Def-GrayAug can be found in Appendix~\ref{sec:uleograyexamples}.
Our \textbf{Exp-Gray} effectively defeats Def-Stan (24.47\% $\rightarrow$ 90.06\%), and our \textbf{Def-Aug-Gray} can still fool Exp-Gray (90.06\% $\rightarrow$ 24.98\%), and also maintain the strength on Exp-Stan (25.98\%).
The ULEO work~\cite{huang2021unlearnable} only explores S1.
}
\vspace{-0.4cm}
\label{fig:fig2}
\end{figure}

\section{Introduction}
\label{sec:int}
The ever-growing amount of easily available online data has enabled the massive progress neural networks have achieved~\cite{schmidhuber2015deep, lecun2015deep}. 
However, online data is often personal or even sensitive, raising concerns about privacy and unauthorized use. 
Several widely in-use data sets have been collected without user consent~\cite{hill2020secretive,birhane2021large} and there is an urgent need for approaches that project users' privacy and allow users to retrain control over their own data.

This need is addressed by \emph{benign data poisoning}, approaches that protect users' data from being used to train a classifier, but do not actively attempt to harm the classifier, as is the case with \emph{malicious data poisoning}.
Key examples are TensorClog~\cite{shen2019tensorclog}, which causes gradient vanishing, and adversarial poisoning~\cite{fowl2021adversarial}, which adds conflicting information to the image data.
Both~\cite{shen2019tensorclog, fowl2021adversarial} mention the usefulness of their approaches for data protection.
Recently,~\cite{huang2021unlearnable} introduced an approach to crafting \emph{UnLearnable Examples (ULEs)}, which has the aim of allowing a \emph{defender} to create examples that are unusable for a \emph{exploiter} to train Deep Neural Networks (DNNs). 
We refer to the specific ULE method proposed by~\cite{huang2021unlearnable} as UnLearnable Examples-Original (ULEO), and call the protected examples it generates ULEOs.
The ULEO method carries out benign data poisoning by learning sample-wise, error-minimizing perturbations that are imperceptible to the human eye.

The goal of this paper is to reveal the road that research should follow in order to understand and improve the state of the art of ULE.
Until now, only adversarial training~\cite{madry2018deep} has been shown to be effective against ULEOs~\cite{huang2021unlearnable,fowl2021adversarial}.
Adversarial training has the disadvantage of being  computationally expensive and it also trades off model accuracy compared to regular training on clean images.
In this paper, we show that a simple pre-filtering method can also effectively defeat ULEOs.
By examining the visual characteristics of ULEO perturbations, as shown in Figure~\ref{fig:fig1}, we observe that they contain few spatial changes but many changes over three color channels.
We refer to such perturbations as channel-wise perturbations.

Building on the insight of Figure~\ref{fig:fig1}, we conjecture that ULEOs mainly exploit channel-wise perturbations.
To test this conjecture, we demonstrate that suppressing channel-wise perturbations will allow classifiers to defeat ULEOs.
We suppress channel-wise perturbations in two ways. 
First, we test simple grayscale pre-filtering as a ULEO countermeasure and find that this \emph{grayscale exploiter} is even more effective than adversarial training in mitigating the effects of ULEOs.
Note that the grayscale exploiter has the same architecture as its corresponding original exploiter but only the input images are pre-filtered to become three (RGB)-channel grayscale images (see Section~\ref{sec:recative} for technical details).
Second, we apply a radical bit-depth reduction~\cite{xu2018feature} and find that this transformation is also quite effective at mitigating ULEOs.

Based on our insight about the importance of channel-wise perturbations, we propose a~\emph{grayscale defender} to generate stronger ULEs by first incorporating grayscale knowledge to prevent channel-wise perturbations, and then promoting spatial patterns via applying standard data augmentations on clean images (see Section~\ref{sec:adapdef} for technical details).
This method is an extension of ULEO that we call ULEO-GrayAugs and that is effective independently of whether grayscale pre-filtering is applied before the classifier is trained.
Further, we show that perturbations generated using simple Multi-Layer Perceptrons (MLPs)~\cite{rumelhart1985learning} do have complex spatial patterns, and are able to fool complex CNNs. 
In this way, we demonstrate the potential benefits of spatial perturbations over channel-wise perturbations.

These observations are the groundwork for further experiments that reveal what we still do not understand about how ULEOs work.  
Specifically, we investigate ULEOs with adversarially trained classifiers and discuss the impact of ULEOs on the adversarial robust accuracy of models, going beyond previous work, which only looks at accuracy on clean images.
We also investigate training on mixed ULEOs and clean data, and moving ULEOs to ImageNet data.
Overall, the aim of our analysis is to provide a better understanding of unlearnable examples and inspire future evaluation in more realistic scenarios.
An overview of our work is shown in Figure~\ref{fig:fig2}.
In sum, this paper makes the following contributions:
\begin{itemize}
    \item We show that ULEOs, the unlearnable examples originally proposed by~\cite{huang2021unlearnable} mainly exploit color channels and that their effects can be mitigated using simple grayscale pre-filtering and without resorting to adversarial training.
    
    \item Building on this insight, we propose ULEO-GrayAugs, which improves upon ULEO by generating grayscale perturbations and using data augmentations for further promoting spatial changes.
    ULEO-GrayAugs is effective against classifiers with and without grayscale pre-filtering. 
    
    \item We, for the first time, generate ULEOs using simple Multi-Layer Perceptrons (MLPs)~\cite{rumelhart1985learning} and find that the resulting perturbations can transfer to CNNs but not the other way around.
    This suggests that CNNs are more vulnerable to ULEOs and generating ULEOs using MLPs serves as an efficient alternative to the current ULEO.
    
    \item We find that the mitigating effect of adversarial training prevents a drop in classifier accuracy both on clean test images and, surprisingly, on adversarial test images. 
    This observations suggests that more work is necessary to achieve generally effective ULEs, and that future work should consider both clean  accuracy and adversarial robust accuracy.

\end{itemize}

\section{Related Work} 

\noindent\textbf{Data Poisoning.} 
The aim of data poisoning can be either permitting/forbidding certain data samples during test time (i.e.$\,$integrity poisoning) or decreasing the general model performance (i.e.$\,$availability poisoning)~\cite{barreno2010security}.
Early studies focus on classical machine learning algorithms, such as linear models and SVMs~\cite{biggio2011support,xiao2015feature,koh2017understanding}, where the poisoning is formulated as a bi-level optimization problem.
Later, data poisoning against deep learning models was also explored.
Backdoor data poisoning~\cite{chen2017targeted,gu2019badnets}, as a type of integrity poisoning, implants specific (trigger) information into a learned model by manipulating training data, in order to cause abnormal model behavior on test samples that have specific triggers. 
In contrast, availability data poisoning aims to degrade a model’s performance by only manipulating the training data~\cite{shen2019tensorclog, huang2021unlearnable, fowl2021adversarial}.
To this end, existing work has relied on gradient vanishing~\cite{shen2019tensorclog}, model error minimization~\cite{huang2021unlearnable} or adversarial examples~\cite{fowl2021adversarial}.
Our work will specifically focus on improving the method of~\cite{huang2021unlearnable}, in terms of both poisoning performance and threat models.

\noindent\textbf{Adversarial Examples.} Adversarial Examples aim to fool models at test time by adding imperceptible perturbations to clean test images~\cite{szegedy2014intriguing,goodfellow2015explaining,carlini2017towards}.
Similar ideas of using grayscale transformation have also used in~\cite{Laidlaw2019functional,zhao2020adversarial} for mitigating color-based adversarial examples.
Adversarial training~\cite{goodfellow2015explaining,madry2018deep} is currently considered the only empirically strong technique for defeating adversarial examples. 
In the context of data poisoning, adversarial training has also been demonstrated to be very effective, i.e., adversarially training a model on poisoned data can secure high enough clean test accuracy~\cite{huang2021unlearnable,fowl2021adversarial}.
However, previous work only focused on the clean test accuracy without exploring the adversarial robustness (i.e.$\,$robust accuracy on adversarial examples around clean test data).
In Section~\ref{sec:adv}, we, for the first time, discuss the impact of ULEs on the adversarial robustness of adversarially trained models.

\noindent\textbf{Privacy Protection.} In addition to data poisoning approaches that can be used to make user data unexploitable during training~\cite{shen2019tensorclog, huang2021unlearnable, fowl2021adversarial, shan2020fawkes}, approaches based on adversarial machine learning have also been developed to protect privacy by misleading the machines during test time, for instance, in person-related recognition~\cite{oh2016faceless, oh2017adversarial, rajabi2021practicality} and social media mining~\cite{larson2018pixel, liu2019pixel}.
Privacy attributes in images were analyzed in depth by~\cite{orekondy2017towards, sattar2020body}.

\section{Threat Model}
\label{sec:thr}
In this section, we describe the threat model we use in this paper.
Following the ULEO work~\cite{huang2021unlearnable}, we introduce two parties: the data defender and exploiter. The defender's goal is to make its uploaded data unlearnable to the exploiter by manipulating them, and the exploiter then trains a classifier from scratch on these uploaded data.
The defender’s success is measured by the accuracy of the classifier on clean test data.
The lower the clean test accuracy, the more successful the defender is considered to be.

\subsection{Defender's and Exploiter's Knowledge}
\label{sec:sce}
We consider the following two scenarios that specify the knowledge of both Defender and Exploiter:

\noindent\textbf{S1: Reactive Exploiter}: The exploiter is aware of ULEs, and so reactively applies specific techniques during model training to mitigate ULEs.

\noindent\textbf{S2: Adaptive Defender}: The defender is aware that the exploiter applies specific techniques for mitigation, and so adapts itself by incorporating the knowledge of such specific techniques into the ULE optimization. 

Specifically, S2 is a new, challenging scenario that has not been explored in the ULEO work~\cite{huang2021unlearnable}.

\subsection{Defender's and Exploiter's Capability}\label{sec:defexpcap}
The defender is only allowed to manipulate the uploaded data but not the training process of the exploiter.
The defender should ensure the stealthiness and maintain the general utility of the manipulated data.
For example, if the data are images posted on some social platform, they should look like normal images.
To ensure stealthiness and maintain general utility, related work has manipulated the image data by adding imperceptible perturbations that are normally restricted by some $L_p$ norm~\cite{huang2021unlearnable,fowl2021adversarial}.
Note that, if the original images are three-channel (RGB) images, the resulting perturbed images need also to be three-channel (RGB) images even with grayscale perturbations.

The exploiter trains its model on collected data until convergence without specific constraints on resources.
The exploiter is assumed to take as input only three-channel (RBG) images.
In this paper, we focus on the case in which the exploiter does not have access to other clean data, and so cannot identify the existence of ULEs in its training data before finally deploying the model.
We also discuss the case in which a clean validation set is available, and point out that in this case, the state-of-the-art adversarial poisoning~\cite{fowl2021adversarial} could be substantially mitigated by early stopping the training process (see Section~\ref{sec:strong} for more details).

\subsection{Unlearnable Example Optimization}
Formally stated, unlearnable examples~\cite{huang2021unlearnable} aim to make a classifier $F$ generalize poorly on the clean image distribution $\mathcal{D}$, from which the clean training set $\mathcal{S}$ is sampled:
\begin{gather}
\label{eq:minmin}
     \max_{\boldsymbol{\delta} \in \Delta} \,\, \mathbb{E}_{(\boldsymbol{x},y) \sim \mathcal{D}} \bigg[ \mathcal{L} \left( F(\boldsymbol{x}; \boldsymbol{\theta}(\boldsymbol{\delta})), y \right) \bigg] \\
     \text{s.t.} \,\, \boldsymbol{\theta}(\boldsymbol{\delta}) = \argmin_{\boldsymbol{\theta}}\sum_{(\boldsymbol{x}_i, y_i) \in \mathcal{S}} \mathcal{L}(F(\boldsymbol{x}_i + \boldsymbol{\delta}_i; \boldsymbol{\theta}), y_i),
\end{gather}
where $\boldsymbol{\theta}$ represents the parameters of the model $F$, and $\mathcal{L}(\cdot;\cdot)$ is the cross-entropy loss, which takes as input a pair of model output $F(\boldsymbol{x}_i; \boldsymbol{\theta})$ and the corresponding label $y_i$.
$\Delta$ denotes the set of the additive perturbations $\boldsymbol{\delta}$.

In order to achieve the above objective, error-minimizing perturbations~\cite{huang2021unlearnable} have been proposed to solve the following min-min bi-level optimization problem:
\begin{gather}\label{eq:ule}
     \argmin_{\boldsymbol{\theta}} \, \mathbb{E}_{(\boldsymbol{x}, y) \in \mathcal{S}} \bigg[ \min_{\boldsymbol{\delta} \in \Delta}\mathcal{L}(F^{'}(\boldsymbol{x}+\boldsymbol{\delta}; \boldsymbol{\theta}), y) \bigg],
\end{gather}
where $F^{'}$ denotes the source model used for perturbation optimization.
This optimization can prevent the classifier $F$ in Eq.~\ref{eq:minmin} from being penalized by the objective function $\mathcal{L}(\cdot;\cdot)$ during training, and as a result can fool the classifier into believing there is ``nothing'' to learn from each perturbed training image $\boldsymbol{x}_i + \boldsymbol{\delta}_i$~\cite{huang2021unlearnable}.
In order to ensure the imperceptibility, the perturbations are normally constrained by $L_{\infty}$ norm $\|\boldsymbol{\delta}\|_{\infty} \leq \epsilon$~\cite{huang2021unlearnable,fowl2021adversarial}.

The inner optimization aims to find the perturbations $\boldsymbol{\delta}$ by minimizing the model’s classification loss, while the outer optimization updates model parameters $\boldsymbol{\theta}$ by training $F^{'}$ on the perturbed images achieved in the inner optimization.
Note that the above inner and outer optimization share the same objective, and the model training starts from scratch in each round while the perturbations accumulate through the whole optimization.
When $F^{'}$ is being updated, $\boldsymbol{\delta}$ is frozen, and vice versa.
The inner and outer optimization will be alternatively implemented and finally terminated when the pre-defined training accuracy on the outer model is met.
The training steps in the outer optimization should be limited compared to standard model training~\cite{franceschi2018bilevel, shaban2019truncated,huang2020metapoison, huang2021unlearnable}.
The detailed pipeline is described in Algorithm~\ref{alg:min-min-perturb} in Appendix~\ref{appendix:algo}.
We focus on the setting of sample-wise perturbations since they are harder to expose in practice and the qualities of ULEOs over other methods are mainly recognized in this setting~\cite{huang2021unlearnable}.

\section{Improving ULEs in Our Two Scenarios}
In this section, we first present how to use grayscale pre-filtering to help the exploiter defeat ULEOs~\cite{huang2021unlearnable}, and then discuss how to improve upon ULEOs against our grayscale exploiter by proposing a grayscale defender.

\subsection{Grayscale Exploiters against ULEOs}\label{sec:recative}
As mentioned before, adversarial training has been recognized to be the only effective technique for defeating unlearnable examples~\cite{huang2021unlearnable, fowl2021adversarial}.
However, it is known to be computationally expensive and also sacrifices the model performance on clean data~\cite{madry2018deep}. 
As we have argued in Section~\ref{sec:int}, ULEO perturbations mainly exploit channel-wise perturbations.
As a result, we propose to use simple grayscale pre-filtering during model training to allow the classifier to mitigate the perturbations of ULEOs.
In this case, the optimization as formulated in Eq.~\ref{eq:minmin} can be modified into: 
\begin{gather}
\label{eq:minmin-gray}
     \argmin_{\boldsymbol{\theta}}\sum_{(\boldsymbol{x}_i, y_i) \in \mathcal{S}} \mathcal{L}(F(Gray(\boldsymbol{x}_i + \boldsymbol{\delta}_i); \boldsymbol{\theta}), y_i),
\end{gather}
where $Gray(\cdot)$ denotes the grayscale filtering\footnote{We implement the grayscale using torchvision \url{https://pytorch.org/vision/stable/_modules/torchvision/transforms/transforms.html\#Grayscale}.} applied on each training input image, $\boldsymbol{x}_i + \boldsymbol{\delta}_i$.
By doing this, the channel-wise changes in the ULEO perturbations are completely removed before the perturbed images are used for model training.
Note that when the classifier is trained on clean data, the gray-scale filtering only leads to small accuracy drop (see Table~\ref{tab:aug}).
This is consistent with previous finding on ImageNet that color information makes little difference to the model accuracy~\cite{xie2018pre}.
The diagram showing the working pipeline can be found in Appendix~\ref{sec:appdiagram}.

\subsection{ULEO-GrayAugs against Grayscale Exploiters}\label{sec:adapdef}
The defender can be adapted to be stronger if it knows that the exploiter has applied gray-scale pre-filtering.
A direct way to achieve this is by incorporating grayscale knowledge into the ULEO optimization.
To this end, we propose ULEO-Gray, which adapts the ULEO optimization in Eq.~\ref{eq:ule} to: 
\begin{gather*}\label{eq:ule-gray}
     \argmin_{\boldsymbol{\theta}} \, \mathbb{E}_{(\boldsymbol{x}, y) \in \mathcal{S}} \bigg[ \min_{\boldsymbol{\delta} \in \Delta}\mathcal{L}(F^{'}(Gray(\boldsymbol{x})+\boldsymbol{\delta}; \boldsymbol{\theta}), y) \bigg],
\end{gather*}
where both $Gray(\boldsymbol{x})$ and $\boldsymbol{\delta}$ still have three (RGB) channels.
Each pixel in $\boldsymbol{\delta}$ is restricted to have the same values in all three channels, in order to prevent the perturbations from exploiting channel-wise changes.

This new optimization can substantially improve upon ULEO against classifiers with grayscale perturbations, but cannot reach comparable performance to that on standard classifiers (see Table~\ref{tab:ulegray} for detailed results).
We argue that it is because the current perturbation optimization only operates on the fixed original image $\boldsymbol{x}$ throughout the whole optimization procedure.
In order to promote ULEO to learn spatial perturbations, we further propose ULEO-GrayAugs, which applies standard (spatial) data augmentations (here, random crop and random horizontal flip) to clean images:
\begin{gather*}\label{eq:ule-grayaug}
     \argmin_{\boldsymbol{\theta}} \, \mathbb{E}_{(\boldsymbol{x}, y) \in \mathcal{S}} \bigg[ \min_{\boldsymbol{\delta} \in \Delta}\mathcal{L}(F^{'}(Gray(Aug(\boldsymbol{x}))+\boldsymbol{\delta}; \boldsymbol{\theta}), y) \bigg],
\end{gather*}
where $Aug(\cdot)$ denotes the data augmentations.
ULEO-GrayAugs allows perturbations to bypass the grayscale pre-filtering and to have more spatial changes.

\section{Experiments}
In this section, we validate the effectiveness of the proposed grayscale pre-filtering in defeating ULEOs, and then demonstrate the strong performance of our ULEO-GrayAugs, which uses grayscale perturbations and data augmentations to allow classifiers to learn spatial changes over channel-wise changes.

\subsection{Experimental Settings}
Following earlier work~\cite{huang2021unlearnable,fowl2021adversarial}, the standard setup has the whole training data consist of either ULEs or clean images. 
In specific experiments we allow the training data to be a mix of clean and ULE training data.
Following~\cite{huang2021unlearnable}, we mainly conduct experiments on CIFAR-10~\cite{krizhevsky2009learning}, which consists of 50000 training images and 10000 test images with the size of 32$\times$32 from 10 classes.
We also discuss ULEs on an ImageNet subset containing the first 100 classes as in~\cite{huang2021unlearnable}.
If not mentioned specifically, we follow~\cite{huang2021unlearnable} to restrict the ULE perturbations by $L_{\infty}$ norm with $\epsilon = 8$ for CIFAR-10 and $\epsilon = 16$ for ImageNet.
\subsection{Grayscale Exploiters against ULEOs}\label{sec:strong}

\begin{figure}[!t]
\centering
\includegraphics[width=0.49\columnwidth]{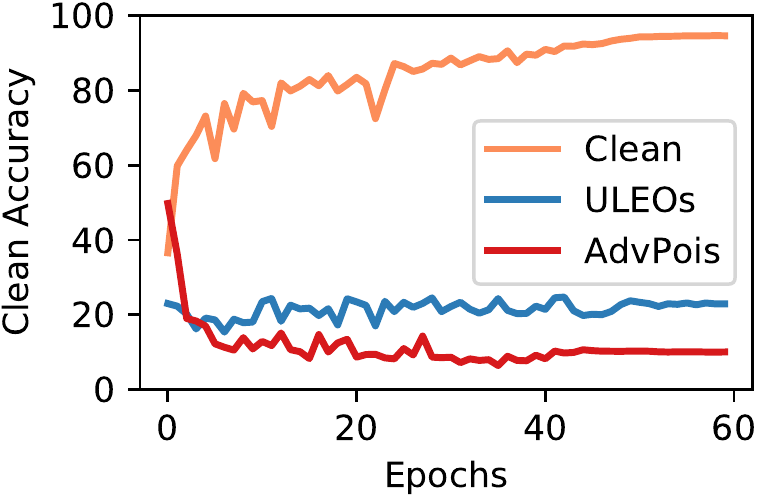}
\includegraphics[width=0.49\columnwidth]{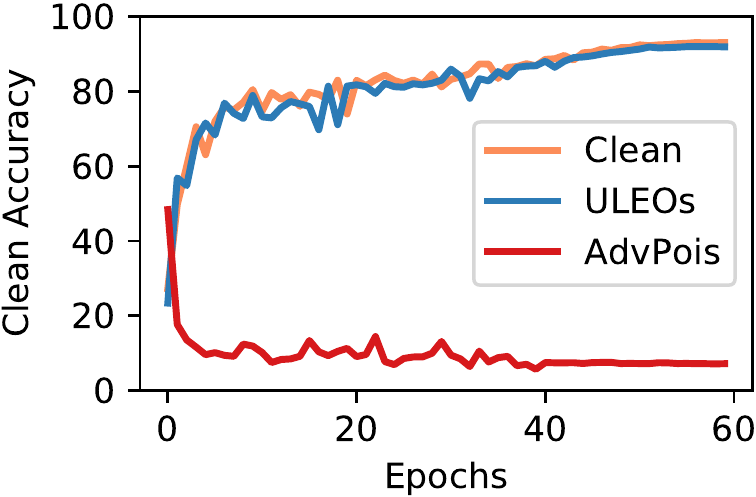}
\vspace{-0.2cm}
\caption{Clean test accuracy (\%) of the standard (left) and grayscale (right) exploiters.
Exploiters are trained on clean data, ULEOs, or adversarial images by the state-of-the-art poisoning method, Adversarial Poisoning (AdvPois)~\cite{fowl2021adversarial}.}
\label{fig:losscurve}
\vspace{-0.3cm}
\end{figure}

\begin{table}[!t]
\caption{Classification accuracy (\%) on CIFAR-10 clean data (for reference) and CIFAR-10 ULEO data mitigated with different methods. Mixup~\cite{zhang2018mixup} (state of the art as per~\cite{huang2021unlearnable,fowl2021adversarial}), Bit-Depth Reduction to 2 bits (BDR-2)~\cite{xu2018feature}, Adversarial Training (AT), and our own grayscale exploiter (Gray), which transforms ULEs to grayscale before classifying. 
 }
\centering\resizebox{0.85\columnwidth}{!}{
\begin{tabular}{l|c|cccccccc}
\toprule[1pt]
Def\textbackslash Exp&w/o&Mixup&BDR-2&AT&Gray\\
\midrule[1pt]
Clean&94.58&93.32&89.12&85.30&93.04\\
ULEOs&24.47&51.01&41.65&84.75&\textbf{90.06}\\

\bottomrule[1pt]
\end{tabular}
}
\vspace{-0.3cm}
\label{tab:aug}
\end{table}

Figure~\ref{fig:losscurve} shows the learning curves of both the standard exploiter and our grayscale exploiter trained on different types of data: clean, ULEOs, and adversarial poisoning examples~\cite{fowl2021adversarial}.
As can be seen, using the proposed grayscale pre-filtering allows the exploiter to effectively mitigate the effects of ULEOs and achieved close performance to the case with clean training data.
We can also observe that Adversarial Poisoning (AdvPois)~\cite{fowl2021adversarial} are effective on both exploiters when only looking at the final accuracy results.
However, it takes multiple training epochs for the model to exploit the perturbations made by AdvPois, making it possible to substantially mitigate AdvPois by early stopping when the exploiter can monitor the model performance on a set of clean validation data (from another reliable source), which is feasible in practice.
The ULEO work~\cite{huang2021unlearnable} also found the same property on random and error-maximizing ULE perturbations.
Note that here we don't discuss TensorClog~\cite{shen2019tensorclog} because it has been shown to be much less effective~\cite{fowl2021adversarial}.

We further compare our grayscale pre-filtering with other techniques for defeating ULEOs.
As can be seen from Table~\ref{tab:aug}, grayscale pre-filtering surpasses the previously thought best pre-filtering technique, Mixup, by 39.05\%.
It also substantially outperforms the current state of the art, adversarial training, and at the same time better maintains the model accuracy when the training data are clean.

It is worth noting that another pre-filtering technique, bit-depth reduction, which can also reduce channel-wise differences, achieves substantial performance against ULEOs.
This result supports our hypothesis that ULEOs mainly exploit channel-wise perturbations.
More detailed results of bit-depth reduction with other bit depths than 2-bit can be found in  Appendix~\ref{sec:exp_appendix}.

\begin{table}[!t]
 \caption{Clean test accuracy (\%) of the standard and grayscale exploiters against ULEs (ULEOs and our ULEOs with grayscale perturbations and/or data augmentations) on CIFAR-10. All results are averaged over 5 runs.}
 \vspace{-0.1cm}
\centering\resizebox{0.75\columnwidth}{!}{
\begin{tabular}{l|cc}
\toprule[1pt]
Def\textbackslash Exp
&w/o Gray&w/ Gray\\
\midrule[1pt]
Clean&94.58&93.04\\
ULEOs&24.47$\pm$4.41&90.06$\pm$3.51\\
ULEO-Augs&23.42$\pm$7.48&91.47$\pm$1.85\\
ULEO-Grays&43.34$\pm$11.49&44.67$\pm$10.47\\
ULEO-GrayAugs&25.98$\pm$7.38&\textbf{24.98$\pm$5.84}\\

\bottomrule[1pt]
\end{tabular}
}
\vspace{-0.2cm}
 \label{tab:ulegray}
\end{table}

\begin{figure}[!t]
\centering
\includegraphics[width=0.49\columnwidth]{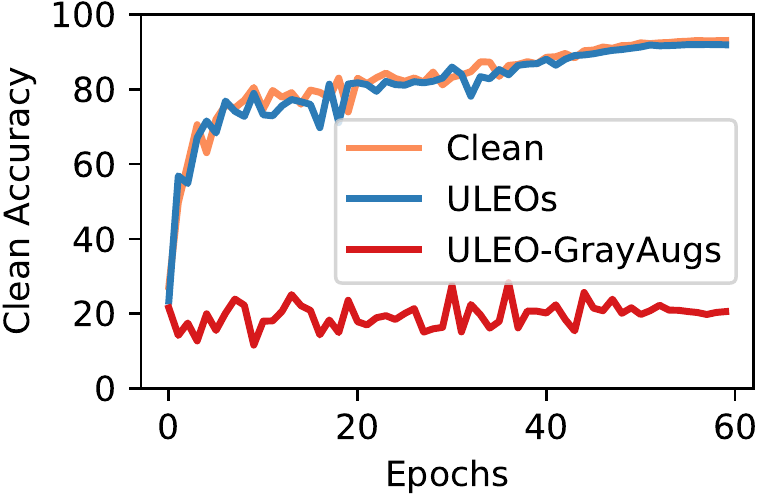}
\vspace{-0.2cm}
\caption{Learning curves of grayscale exploiters trained on clean data, ULEOs, or ULEO-GrayAugs.}
\label{fig:losscurve2}
\vspace{-0.5cm}
\end{figure}

\subsection{ULEO-GrayAugs against Grayscale Exploiters}
As we have shown that the ULEOs are vulnerable to grayscale exploiters, here we demonstrate the effectiveness of our ULEO-GrayAugs, which improve upon ULEOs by generating perturbations with a larger spatial variation.
As can be seen from Table~\ref{tab:ulegray}, our ULEO-GrayAugs still yield a low accuracy (24.98\%) on the grayscale exploiter.
This confirms ULEO-GrayAugs exploit spatial perturbations to bypass the grayscale pre-filtering.
More detailed evidence with the training curves of ULEOs vs. ULEO-GrayAugs can be found in Figure~\ref{fig:losscurve2}.
More interestingly, our ULEO-GrayAugs method also yields similarly good performance on the standard exploiter, making it a generally stronger ULE solution than the ULEO method.
We also find that solely using data augmentations (ULEO-Augs) will make no difference from ULEOs and solely using grayscale (ULEO-Grays) also does not lead to optimal results.

The above experiments are based on the assumption that the defender can use the same model as the exploiter to generate ULEs. However, it might not be the case in practical scenarios with unknown target models.
To test how the effectiveness of ULEO-GrayAugs generalizes to practical scenarios, we test their cross-model transferability.
As can be seen from Table~\ref{tab:trans}, ULEO-GrayAugs can be successfully generated independently of the model architectures.
In addition, they well maintain their strong effects when being transferred from one model to another.

\begin{table}[!t]
         \caption{Clean test accuracy (\%) of exploiters in the columns when defenders choose a model in the rows for generating ULEO-GrayAugs. We consider three architectures: ResNet-18 (RN-18), DenseNet-121 (DN-121), and VGGNet-11 (VN-11), and report results for standard/grayscale exploiters.}
         \vspace{-0.1cm}
\renewcommand{\arraystretch}{1}
      \centering
      \resizebox{0.85\columnwidth}{!}{
        \begin{tabular}{l|ccccc}
\toprule[1pt]
Def\textbackslash Exp&RN-18&DN-121&VN-11\\
\midrule
RN-18&28.35/28.97&25.13/23.99&33.07/28.78\\
DN-121&28.85/30.01&28.94/25.79&30.80/25.00\\
VN-11&11.47/11.14&14.28/13.26&14.50/13.23\\

\bottomrule[1pt]
\end{tabular}}
\vspace{-0.0cm}
\label{tab:trans}
\end{table}

\begin{table}[!t]
    \caption{Clean test accuracy (\%) of a model trained by the exploiter (with or without grayscale pre-filtering) on an original (Ori) training set containing 5\% of the official training data of CIFAR-10 with additional (Add) training data of Clean, ULEOs, or ULEO-GrayAugs. 
    For each case with all training data being clean, we directly report the model accuracy, while for each case with partial clean and ULEs, we report its accuracy difference from the case with the same amount of clean training data. Note that after adding 30\% the model accuracy starts getting saturated. Similar results have been found for the cases with Ori as 10\% (see Table~\ref{tab:perc_appendix} in Appendix~\ref{sec:exp_appendix}).}
    \vspace{-0.1cm}
    \centering\resizebox{0.95\columnwidth}{!}{
        \begin{tabular}{cc|c|ccc}
            \toprule[1pt]
            Ori& Add & Gray & Clean & ULEOs & ULEO-GrayAugs  \\
            \midrule[1pt]

            5\% & - & $\times$ & 63.89 & - & - \\
            5\% & 10\% & $\times$ & 80.83 & \textbf{-13.55} & -4.61 \\
            5\% & 30\% & $\times$ & 89.26 & \textbf{-22.12} & -10.48 \\
            5\% & - & $\checkmark$ & 61.99 & - & - \\
            5\% & 10\% & $\checkmark$ & 81.61 & -1.54 & \textbf{-6.11} \\
            5\% & 30\% & $\checkmark$ & 87.95 & -0.54 & \textbf{-15.84} \\
            \bottomrule[1pt]
        \end{tabular}}
        \vspace{-0.2cm}
    \label{tab:perc}
\end{table}

\subsection{ULEs Mixed with Clean Training Data}

So far, our experiments have been conducted in the setting that the whole training data are ULEs or clean images.
However, it is realistic that exploiters may have access to other clean training data from another source.
Here we test the effectiveness of ULEs (ULEOs and our ULEO-GrayAugs) when different proportions of clean and perturbed
data are used to train the classifier.
In this case, ULEs are considered effective if adding them leads to significantly less increased accuracy compared with adding the same amount of clean data.

Table~\ref{tab:perc} compares the accuracy before and after adding clean data or ULEs to the original training set.
As can be seen, in all cases, adding extra data improves model accuracy.
This is consistent with previous findings~\cite{huang2021unlearnable,fowl2021adversarial}.
However, compared with adding clean data, adding the same amount of ULEs leads to less improvement.
Specifically, for standard exploiters, ULEOs are stronger and ULEO-GrayAugs also yield substantial accuracy drop.
However, for grayscale exploiters, ULEO-GrayAugs are stronger but ULEOs yield little accuracy drop.

\begin{table}[!t]
 \caption{Clean test accuracy (\%) of standard and grayscale exploiters against ULEs (ULEOs and our ULEO-GrayAugs) on ImageNet subset ($\epsilon$ = 16). The perturbations were class-wise instead of sample-wise.}
 \vspace{-0.1cm}
\centering\resizebox{0.7\columnwidth}{!}{
\begin{tabular}{l|cc}
\toprule[1pt]
Def\textbackslash Exp
&w/o Gray&w/ Gray\\
\midrule[1pt]
Clean& 59.66&58.18\\
ULEOs&4.88&8.52\\
ULEO-GrayAugs
&7.04&  6.46\\

\bottomrule[1pt]
\end{tabular}
}
\vspace{-0.4cm}
 \label{tab:acc_imagenet_nojitter}
\end{table}

\subsection{ULEs on ImageNet}
\label{sec:ima}
In order to verify the generalizability of ULEs (ULEOs and our ULEO-GrayAugs) to larger datasets, we test on ImageNet~\cite{deng2009imagenet} images. Table~\ref{tab:acc_imagenet_nojitter} shows the results achieved by directly using the official training code~\footnote{\url{https://github.com/HanxunH/Unlearnable-Examples}} of~\cite{huang2021unlearnable}.
Note that we remove the ``color jitter'' data augmentation during the model training because it naturally affects channel-wise perturbations, making the usefulness of grayscale pre-filerting difficult to validate. 
As can be seen, the improvement of our grayscale exploiter and ULEO-GrayAugs still holds.
However, the improvement is limited compared to the reported results on CIFAR-10.
By carefully checking the official code, we identify an implementation error\footnote{In personal communication, the authors acknowledge the error and are planning to address it in order to support further research.} that mistakenly trains classifiers on (incorrect) class-wise perturbations~\cite{huang2021unlearnable}, which make ULEOs achieve strong results much more easily on exploiters with/without grayscale pre-filtering.
After clearing up this error, we find that the model training part of the bi-level ULEO optimization cannot converge to a pre-defined high accuracy after a small-scale hyper-parameter search.
We suspect it might be because ImageNet has more classes and images with higher visual complexity, while having less training samples per class.
For this reason, we also try generating ULEOs on ImageNet with only 10 classes and find that the optimization can successfully converge, but the resulting perturbations have very limited effects.
The perturbations for both experiments are shown in Appendix~\ref{sec:cifarlarge}.
There, we also visualize perturbations generated for upsampled (224$\times$224) CIFAR-10 images to show that increasing the image size alone does not change the channel-wise patterns of perturbations.
We leave further exploration for future work.

\section{Additional Practical Insights into ULEs}\label{sec:addi}

In this section, we provide additional practical insights into ULEs that have not been discussed in previous work~\cite{huang2021unlearnable,fowl2021adversarial}.
We discuss generating ULEs while using MLPs to fool CNNs from the defender perspective. We also explore from the exploiter perspective the effectiveness of adversarial training in maintaining adversarial robust accuracy of models against ULEs.

\begin{figure}[!t]
\centering
\includegraphics[width=0.75\columnwidth]{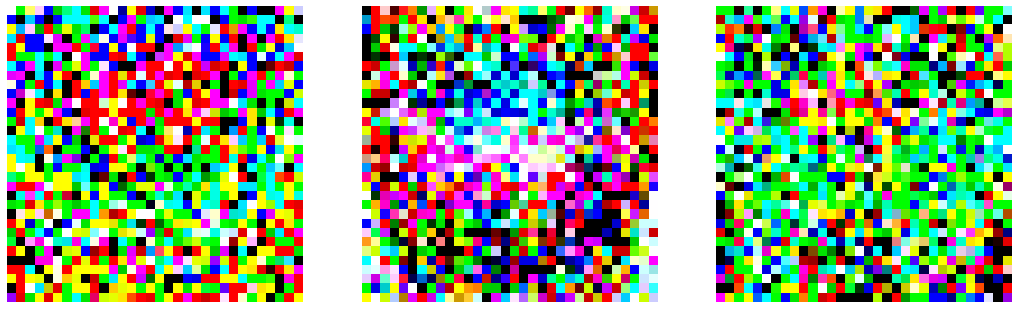}
\caption{Perturbations of ULEOs generated using MLPs for the same three CIFAR-10 images shown in Figure~\ref{fig:fig1}. }
\vspace{-0.1cm}
\label{fig:noiseMLP}
\end{figure}

\begin{table}[!t]
 \caption{Clean test accuracy (\%) of MLP vs. CNNs that are trained on clean images/MLP-generated ULEOs.}
 \vspace{-0.1cm}
\centering\resizebox{\columnwidth}{!}{
\begin{tabular}{l|cccc}
\toprule[1pt]
Def\textbackslash Exp&MLP&RN-18&DN-121&VN-11\\
\midrule[1pt]
MLP&48.66/47.50&94.58/21.55&95.19/15.84&91.66/81.88\\

\bottomrule[1pt]
\end{tabular}
}
\vspace{-0.5cm}
\label{tab:mlp}
\end{table}

\subsection{ULEs on MLPs}

The exploration of ULEOs in the original~\cite{huang2021unlearnable} and our work has so far been limited to Deep Neural Networks (CNNs).
However, it remains unclear whether the underlying mechanism of the ULEOs are specific to CNNs or also generalizable to other machine learning algorithms.  
To address this concern, we also try generating ULEOs on a simple machine learning algorithm, Multi-Layer Perceptrons (MLPs)~\cite{rumelhart1985learning}.
Specifically, we use flattened ULEOs or clean images as inputs to MLPs.

As can be seen from Table~\ref{tab:mlp}, the ULEOs generated using MLPs are not able to succeed against MLP exploiters.
However, surprisingly, they can be very strong against CNN exploiters, especially on ResNet-18 and DenseNet-121.
This observation provides a generally more efficient way to generate successful ULEOs on CNNs, i.e.$\,$using a simple MLP as the defender.
A closer look at the training curves (see Figure~\ref{fig:cureveMLP} in Appendix~\ref{sec:exp_appendix}) of the CNNs on MLP-generated ULEOs suggests that MLP-generated ULEOs, similar to those generated by AdvPois~\cite{fowl2021adversarial}, can also be substantially mitigated by early stopping.
We further try the reverse direction by transferring ULEOs from CNNs to MLPs, but find that it does not work.
This suggests that CNNs are generally much more vulnerable to ULEOs than MLPs.

In order to shed light on the specific properties of ULEOs on MLPs vs. CNNs, we visualize the perturbations of the ULEOs generated on MLPs in Figure~\ref{fig:noiseMLP}.
As can be seen, different from the perturbations of ULEOs that are generated using CNNs (as shown in Figure~\ref{fig:fig1}), these perturbations on MLPs naturally contain many spatial changes. 
This difference suggests that the underlying working principles of these two types of ULEOs might be different.
We leave more detailed exploration of this difference for future work.

\begin{table}[!t]
 \caption{Clean and adversarial robust (against FGSM and PGD with 20 steps) test accuracy (\%) of exploiter that is adversarially trained on clean images, ULEOs, or our ULEO-GrayAugs. All models are adversarially trained using PGD with 7 steps.}
 \vspace{-0.2cm}
\centering\resizebox{0.75\columnwidth}{!}{
\begin{tabular}{l|c|cc}
\toprule[1pt]
AT-\textbackslash Test&Clean&FGSM&PGD\\
\midrule[1pt]
Clean&85.30&54.01&46.59\\
ULEOs&84.75&52.42&40.34\\
ULEO-GrayAugs&84.38&53.07&43.80\\

\bottomrule[1pt]
\end{tabular}
}
\vspace{-0.5cm}
 \label{tab:adv}
\end{table}

\subsection{ULEs Meet Adversarial Examples}
\label{sec:adv}
Current work has demonstrated that Adversarially Trained models on ULEs (AT-ULE models) can still achieve high clean test accuracy~\cite{huang2021unlearnable,fowl2021adversarial}.
This is explainable because the AT-ULE model is expected to be effective in maintaining normal model behaviors on any test examples that fall into the $\epsilon$-ball of ULEs.
Since clean test examples obviously fall into this ball, clean test accuracy of AT-ULE model should be comparable to the training accuracy.

Despite the above observation that adversarial training prevents ULEs from decreasing the clean test accuracy, we find that it can also surprisingly, prevents the drop of robust test accuracy. 
It is still unclear if it prevents the drop of adversarial robust accuracy (i.e., accuracy on adversarial examples around clean test data).
This is related to a realistic scenario in which an exploiter has implemented computationally expensive adversarial training for achieving high robust accuracy around clean data but might not know its training data are actually ULEs (rather than clean data). 
In this scenario, the resulting AT-ULE models would not be intended to achieve high adversarial robust accuracy around clean test data because the AT-ULE models have been only trained on ULEs and expected to secure the adversarial robust accuracy around ULE test data.

However, as can be seen from Table~\ref{tab:adv}, AT-ULE models trained on the current ULE methods (ULEOs and ULEO-GrayAugs) only yield slightly lower robust accuracy than that of the AT models trained on clean data.
This suggests that in practice, the current ULE methods are still vulnerable to adversarial training in terms of clean test accuracy, but also, adversarial robust accuracy.
More details about this unexpected phenomenon are worth exploring by future research, in order to achieve comprehensively stronger ULEs against adversarial training.

\section{Discussion}\label{sec:dis}
In this section, we first provide general insight into understanding ULEs.
Then we discuss limitations of our work and potential ways to address them in future work. 
\subsection{ULEO Principles}
In this paper, we have shown ULEOs mainly exploit simple, channel-wise perturbations.
However, the reason why ULEOs have this ``lazy'' behavior remains unclear.
In the following, we provide two related perspectives. 

\noindent\textbf{Model architecture matters.} We have shown that generating ULEOs on MLPs do meet our expectation of exploiting complex, spatial perturbations over simple, channel-wise perturbations, and these perturbations are also effective against CNNs. 
This implies that the ``lazy'' behavior might be related to the fact that CNNs tend to use shortcut features~\cite{geirhos2020shortcut}.
Since convolutions have a channel dependency and have the translation-invariant property, exploiting channel-wise perturbations could be simpler than exploiting more complex, spatial perturbations. 

\noindent\textbf{Lack of data may not be relevant.} If lack of data is the reason for the ``lazy'' behavior, we would expect that by using data augmentations, this behavior will be alleviated.
However, our results suggest that the ULEO-Augs, which incorporate data augmentations during optimization, still exploit channel-wise perturbations, as shown in Figure~\ref{fig:pert_uleos-augs} of Appendix~\ref{sec:exp_appendix}.
Also, ULEO-Augs have been shown to achieve almost the same performance as ULEOs.
Our results of ULEO-GrayAugs have shown that data augmentations might play a role on top of grayscale perturbations.
Specifically, we find that ULEO-GrayAugs promote the use of complex, spatial perturbations (see Figure~\ref{fig:fig2}) and lead to much stronger results than ULEOs.

\subsection{Limitations and Outlook}
\label{sec:lim}
In this section, we sketch the road ahead for research on unlearnable examples by describing the remaining limitations of our investigation and providing an outlook onto important open points that should be addressed by related work.

\noindent\textbf{Threat model.}
We have seen that extending the threat model to include more realistic scenarios yields deeper insight into ULEs.
Future work should continue to expand the threat model.
First, the threat model should explicitly specify the constraints on available resources.
For the exploiter, adversarial training may be computationally too expensive to be a viable solution.
For the defender, ULE generation must be simple and efficient so that it remains feasible to as many users as possible to apply in order to protect their data.
We have demonstrated promising results for ULEs generated using MLPs, a simple model, but more research is necessary to determine which other simple models are also effective and to understand why.

The threat model should also explicitly specify that the intent of the defender is benign.
Our experiments show that ULEs are not particularly dangerous. 
Clean data is more useful to the exploiter than ULEs (ULEOs or ULEO-GrayAugs), but adding ULEs does not hurt the model and may even improve performance.
Future ULEs must continue to provide users with the confidence that their ULE-protected data will simply fail to improve the classifier, rather than negatively impacting its performance.
Users with the intent to protect their own data do not want to be mistaken for malicious attackers, resulting in e.g., being banned from the platform (or worse) for attempting to actively harm the performance of a classifier.

\noindent\textbf{Adversarial Training.}
Our work has shown that adversarial training mitigates the effect of ULEs (both ULEOs and our ULEO-GrayAugs) and the resulting classifiers are unexpectedly robust to adversarial examples around clean data.
Moving forward, researchers should measure the strength of ULEs against adversarially trained models on both original data and adversarial examples.
Future work should seek to understand this surprising finding, and strive for ULEs that remain effective in the face of adversarial training.

\noindent\textbf{Discouraging Shortcuts.} The strength of ULEO perturbations, as discussed above, might be related to the extent to which ULEO approaches exploit shortcut features. 
Future work should further explore limiting the possibilities of shortcuts during ULE generation. 
Such research promises to lead to more robust data protection and also better theoretical understanding of ULEs.

\noindent\textbf{Data variety.} We have carried out the evaluation with data comparable to that used by~\cite{huang2021unlearnable}.
Future work on ULEs should test on a wider range of data sets.
First, larger data sets with more training data and/or more competing classes could be relevant.
Second, data sets with very high or very low visual complexity could also yield additional insight into how and why ULEs work.
\section{Conclusion}
\label{sec:con}
In this paper, we have advanced the state of the art by extending unlearnable examples as originally studied by~\cite{huang2021unlearnable} (ULEOs) to a more realistic scenario including a grayscale exploiter.
This scenario revealed the dependence of ULEOs on color perturbations and led us to propose ULEO-GrayAugs, a novel version of ULEO that remains robust under the more realistic scenario.
We demonstrate, for the first time, that ULEOs generated using simple MLPs can have strong impact on CNN-based classifiers and provide an efficient way to generate successful ULEOs.
We also present evidence, from tests of adversarial robust accuracy, that ULEO-GrayAugs has future room for improvement.
We have also discussed several directions of future work, for which our paper lays a foundation.

\section*{Acknowledgments}
This work was carried out on the Dutch national e-infrastructure with the support of SURF Cooperative.


{\small
\bibliographystyle{ieee_fullname}
\bibliography{egbib}
}

\clearpage
\newpage

\appendix
\section{Additional Experimental Results}
\label{sec:exp_appendix}

\begin{table}[h]
\caption{Clean test accuracy (\%) on CIFAR-10 with clean training data (for reference) or other poisoned training data (ULEOs, AdvPois, or our ULEO-GrayAugs) mitigated with Bit-Depth Reduction to different bits~\cite{xu2018feature}.}
\renewcommand{\arraystretch}{1}
      \centering
      \resizebox{0.8\columnwidth}{!}{
        \begin{tabular}{l|ccccc}
\toprule[1pt]
&w/o&BDR-2&BDR-4&BDR-6\\
\midrule
Clean&94.58&89.12&93.98&94.48\\
ULEOs&24.40& 47.10&22.92&21.39\\
\bottomrule[1pt]
\end{tabular}}
\label{tab:bdr_appendix}
\end{table}

\begin{table}[h]
    \caption{Clean test accuracy (\%) of a model trained by the exploiter (with or without grayscale) on an original (Ori) training set containing 10\% of the official training data of CIFAR-10 plus additional (Add) training data of Clean, ULEOs, or ULEO-GrayAugs. 
    For each case with all training data being clean, we directly report the model accuracy, while for each case with partial clean and ULEs, we report its accuracy difference from the case with the same amount of clean training data. Note that after adding 30\% the model accuracy starts getting saturated.}
    \centering\resizebox{\columnwidth}{!}{
        \begin{tabular}{cc|c|ccc}
            \toprule[1pt]
            Ori& Add & Gray & Clean & ULEOs & ULEO-GrayAugs  \\
            \midrule[1pt]
            10\% & - & $\times$ & 73.48 & - & - \\
            10\% & 10\% & $\times$ & 83.30 & \textbf{-3.64} & -1.23 \\
            10\% & 30\% & $\times$ & 89.85 & \textbf{-11.22} & -5.61 \\
            10\% & -& $\checkmark$ & 74.23 & - & - \\
            10\% & 10\% & $\checkmark$ & 82.72 & -0.05 & \textbf{-2.30} \\
            10\% & 30\% & $\checkmark$ & 88.46 & -0.18 & \textbf{-7.56}\\
            
            \bottomrule[1pt]
        \end{tabular}}
    \label{tab:perc_appendix}
\end{table}

\begin{figure}[h]
\centering
\includegraphics[width=0.7\columnwidth]{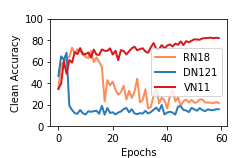}
\caption{Learning curves of the CNNs trained on ULEOs generated using MLPs.}
\label{fig:cureveMLP}
\end{figure}

\begin{figure}[h]
    \centering
    \includegraphics[width=0.75\linewidth]{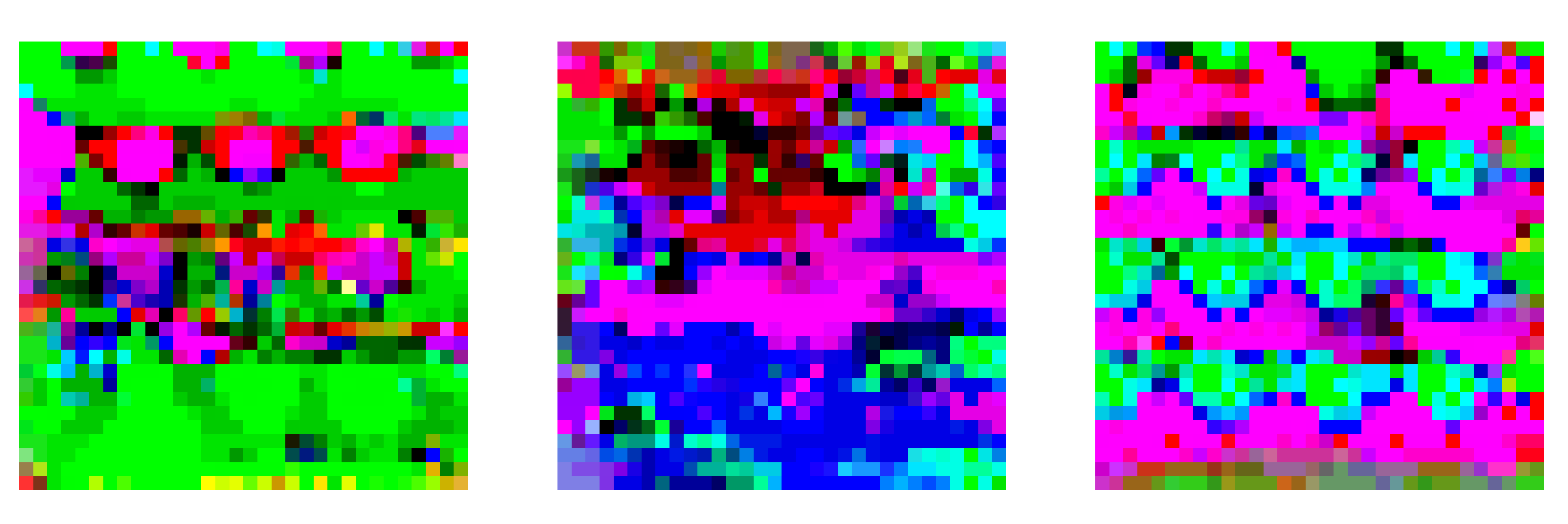}
    \caption{Perturbations generated with only data augmentations (ULEO-Augs) for the three clean images in Figure~\ref{fig:fig1}. They still exploit little spatial patterns as perturbations of ULEOs.}
    \label{fig:pert_uleos-augs}
\end{figure}

\newpage

\section{Implementation details}\label{appendix:algo}

\begin{algorithm}
    \caption{Sample-wise error-minimizing perturbations}
    \label{alg:min-min-perturb}
    \begin{algorithmic}[1]
    \State {\bfseries Input:} Initialized model weights $\boldsymbol{\theta}_{0}$, random perturbations $\boldsymbol{\delta}_0$, $\mathcal{L}_{\infty}$ perturbation constraints $\epsilon$, clean training data ($\boldsymbol{x}$, y) $\in \mathcal{S}$, stop error $\lambda$, model training epochs $M$, iteration count $i$.
    \State {\bfseries Output:} $\boldsymbol \delta$
    \State $i = 0$
    \Repeat
        \For{$m$ {\bfseries in} $1 \cdots M$}
            \State $\boldsymbol{\theta}_{i + 1} \leftarrow$ Optimize($\boldsymbol{x}+{\boldsymbol{\delta}_i, y}, \boldsymbol{\theta}_{i}$) \Comment{Eq.~\ref{eq:ule} Outer}
        \EndFor
            \State $\boldsymbol{\delta}_{i + 1} \leftarrow $ $\min (\mathcal{L}(F^{'}(\boldsymbol{x}+\boldsymbol{\delta}_i; \boldsymbol{\theta}_{i + 1}), y))$\Comment{Eq.~\ref{eq:ule} Inner}
            \State $\boldsymbol{\delta}_{i + 1} \leftarrow$ Clip($\boldsymbol{\delta}_{i + 1}, -\epsilon, \epsilon$)
        \State Error $\leftarrow$ Eval($F^{'}(\boldsymbol{x} + \boldsymbol{\delta}_{i+1}; \boldsymbol{\theta}_{i + 1}), y))$
        \State $i \leftarrow i + 1$
    \Until{Error$< \lambda$} \\
    \Return $\boldsymbol \delta$
    \end{algorithmic}
\end{algorithm}

\section{ULEO Pipeline}\label{sec:appdiagram}
\begin{figure}[h]
\centering
\includegraphics[width=\linewidth]{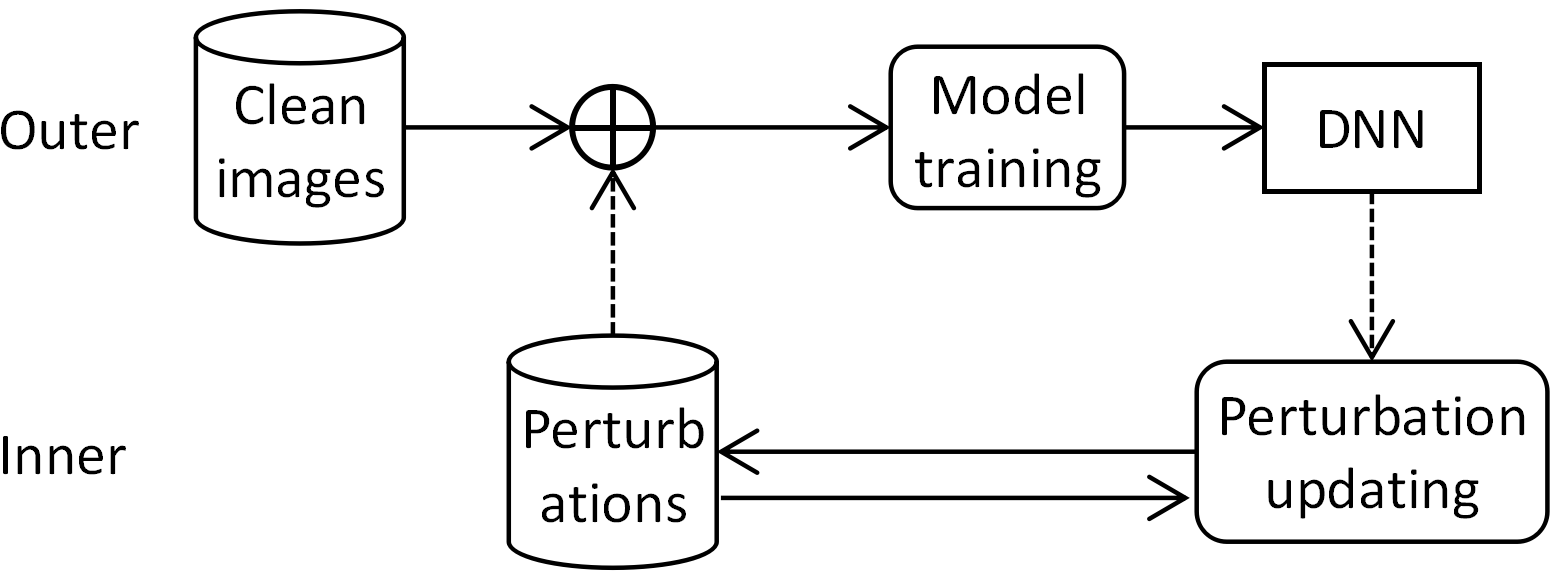}
\caption{The pipeline of bi-level optimization for generating ULEOs. The dashed lines represent the information exchange between inner and outer optimization.}
\label{fig:diagram}
\end{figure}

The pipeline of bi-level optimization for generating ULEOs is described in Figure~\ref{fig:diagram}.
First, as shown in the first row, the perturbations, which are randomly initialized, are added to clean images, and the resulting perturbed images are used to train a DNN model for multiple epochs in the outer optimization.
Then, as shown in the second row, given the trained DNN model from the outer optimization, perturbations are optimized for multiple iterations.
Finally, the optimized perturbations are added to clean images for model training in the next round of outer optimization.
The above whole process is repeated until the pre-defined training accuracy is met.

\section{Additional Examples of ULEOs}\label{sec:uleoexamples}

\noindent\begin{minipage}{\textwidth}
\centering
\includegraphics[width=0.37\linewidth]{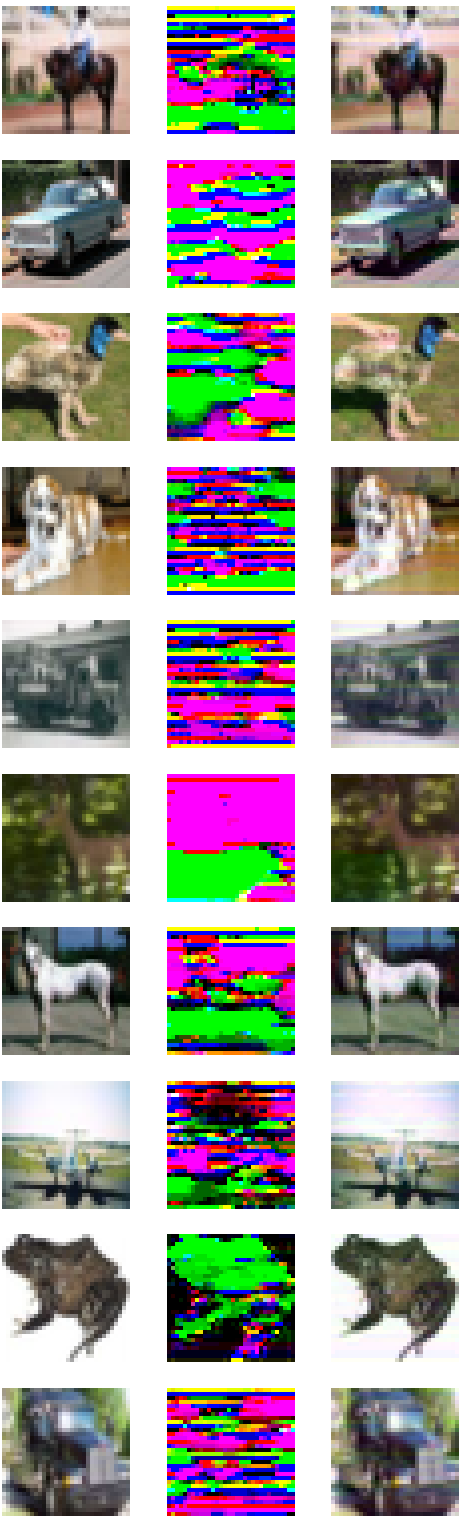}
\hspace{1cm}
\includegraphics[width=0.37\linewidth]{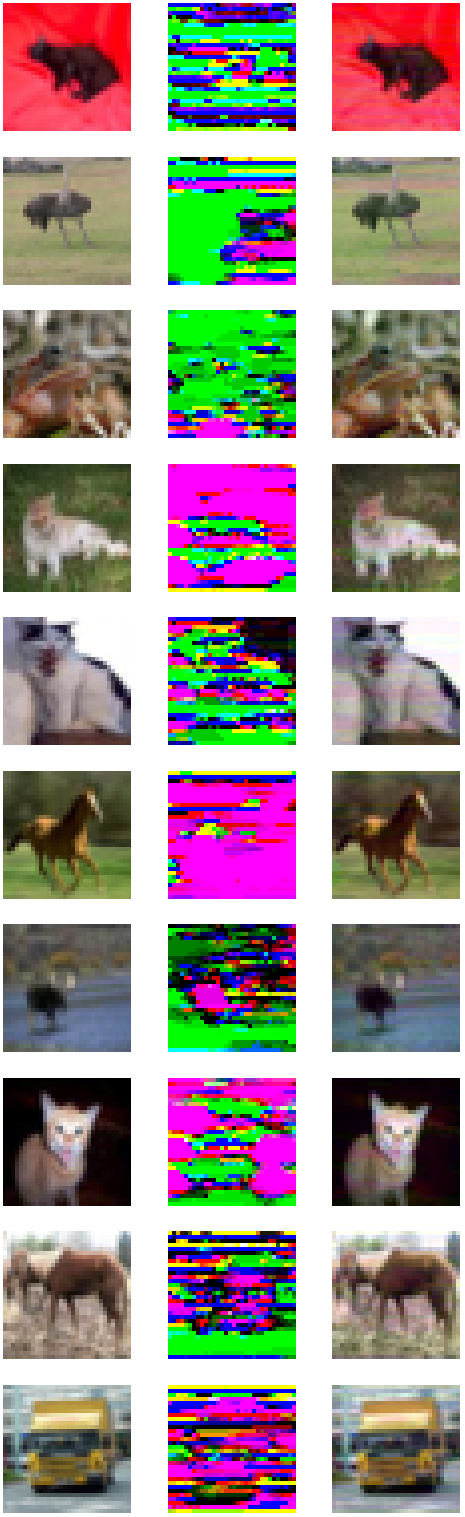}
\captionof{figure}{Additional ULEO Examples on CIFAR-10: Original images (left), ULEO perturbations (middle), and perturbed images (right).}
\label{fig:moreuleo}
\end{minipage}

\section{Additional Examples of ULEO-GrayAugs}\label{sec:uleograyexamples}

\noindent\begin{minipage}{\textwidth}
\centering
\includegraphics[width=0.37\linewidth]{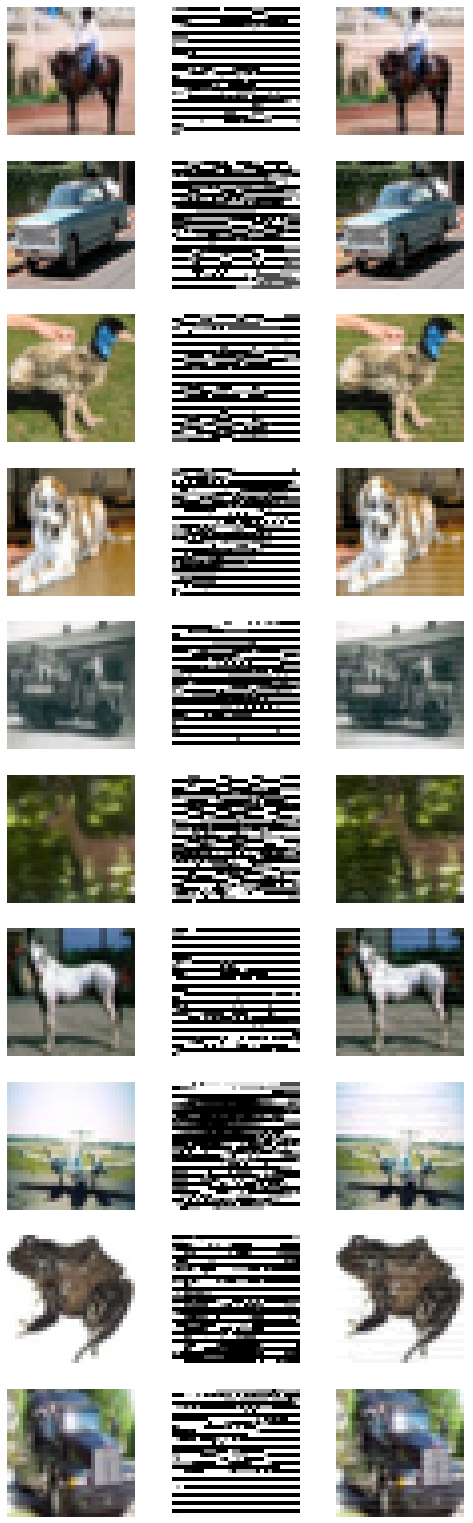}
\hspace{1cm}
\includegraphics[width=0.37\linewidth]{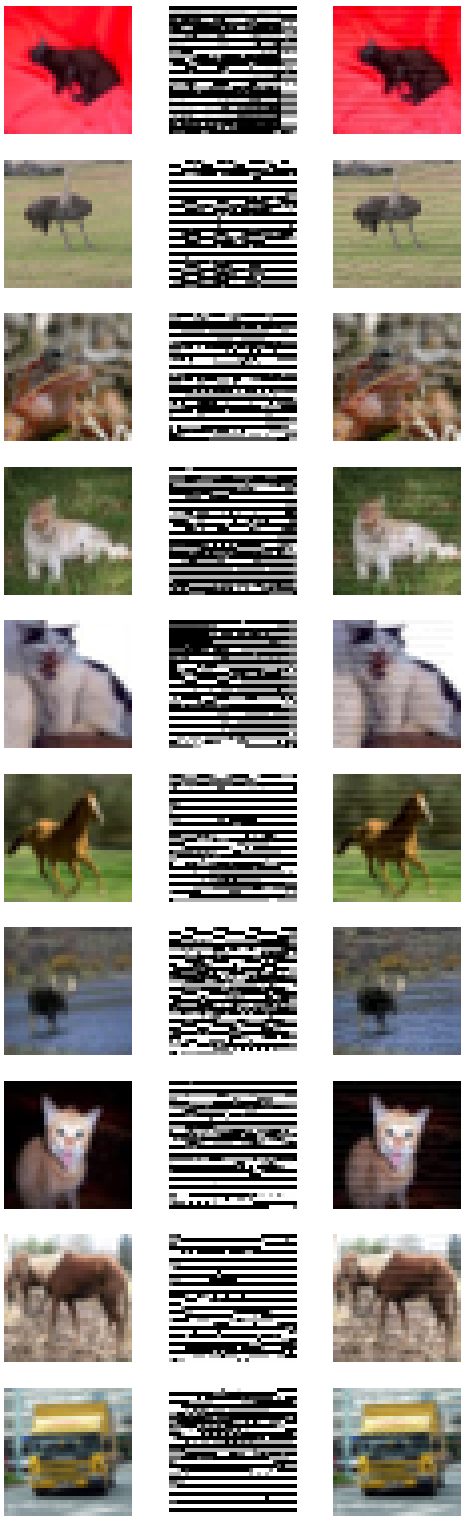}
\captionof{figure}{Additional ULEO-GrayAug Examples on CIFAR-10: Original images (left), ULEO-GrayAug perturbations (middle), and perturbed images (right). Examples for the same clean images as in Figure~\ref{fig:moreuleo} are shown.}
\label{fig:moreuleograyaug}
\end{minipage}

\clearpage

\section{ULEOs on ImageNet (100 and 10 classes) and on Upsampled CIFAR-10}\label{sec:cifarlarge}

\noindent\begin{minipage}{\textwidth}
\centering
\includegraphics[width=0.75\linewidth]{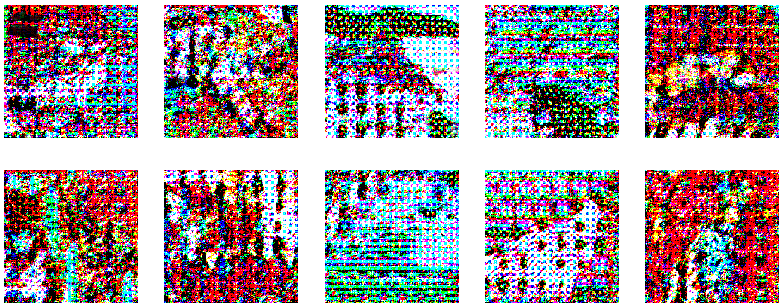}
\captionof{figure}{Class-wise ULEO Perturbations (224 $\times$ 224) on ImageNet subset with 100 classes used for the (incorrect) experiments reported in Table~\ref{tab:acc_imagenet_nojitter}.}
\label{fig:imn100}
\vspace{1cm}
\end{minipage}

\noindent\begin{minipage}{\textwidth}
\centering
\includegraphics[width=0.45\linewidth]{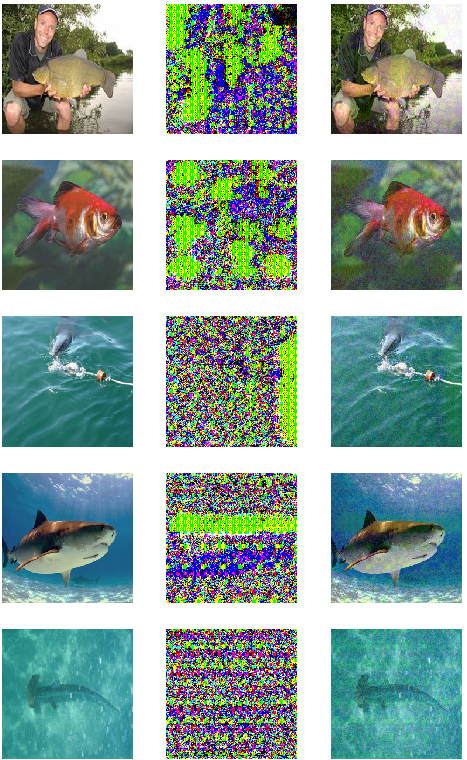}
\hspace{1cm}
\includegraphics[width=0.45\linewidth]{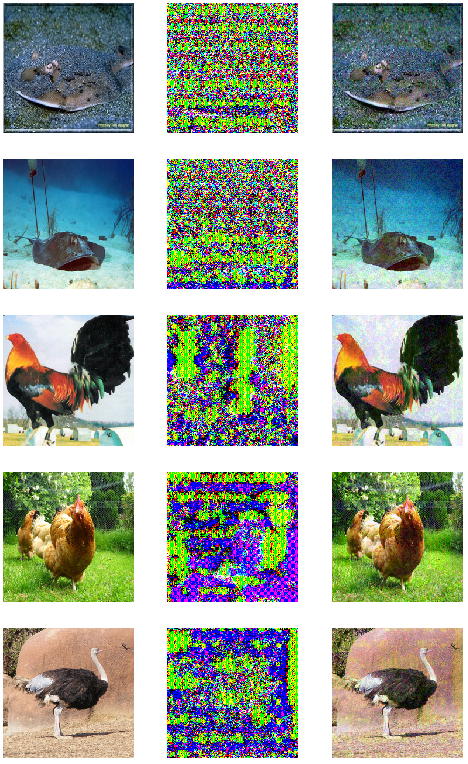}
\captionof{figure}{ULEO Examples (224 $\times$ 224) on ImageNet subset with 10 classes: Original images (left), ULEO perturbations (middle), and perturbed images (right).}
\label{fig:imn10}
\end{minipage}

\clearpage
\noindent\begin{minipage}{\textwidth}
\centering
\includegraphics[width=0.45\linewidth]{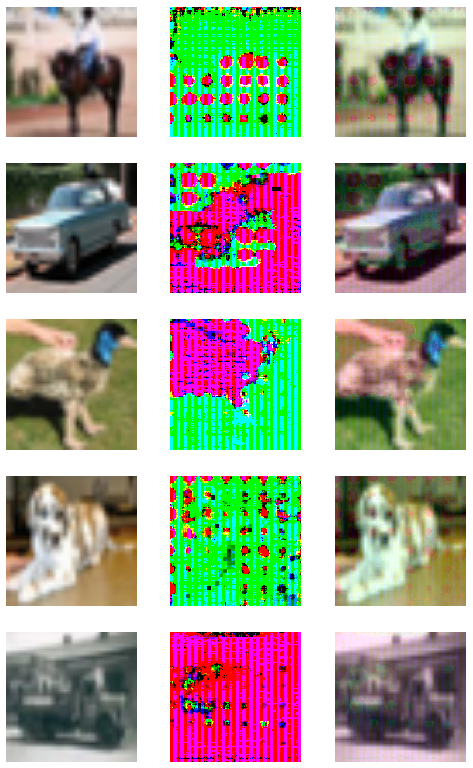}
\hspace{1cm}
\includegraphics[width=0.45\linewidth]{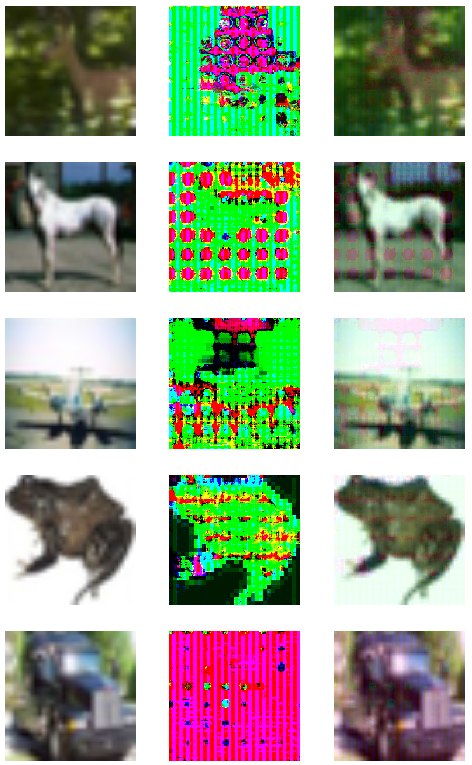}
\captionof{figure}{ULEO Examples on upsampled CIFAR-10 (224 $\times$ 224): Original images (left), ULEO perturbations (middle), and perturbed images (right).}
\label{fig:cifarlarge}
\end{minipage}

\end{document}